\pdfoutput=1
\documentclass{article}
\usepackage{cite,spconf,amsmath,graphicx,enumitem,amssymb,color,bm,subfigure}
\usepackage{float}
\usepackage{mathrsfs}
\usepackage{amsfonts}
\usepackage{amsthm}
\usepackage{indentfirst}
\usepackage{array}
\usepackage{epstopdf}
\usepackage{cite}
\usepackage{enumerate}
\usepackage{bm}
\usepackage[linesnumbered,ruled,vlined]{algorithm2e}
\usepackage{setspace}
\usepackage{multirow}
\usepackage{verbatim}
\usepackage{stfloats}
\usepackage{color}
\usepackage{makecell}
\usepackage{url}

\title{WITT: A Wireless Image Transmission Transformer For Semantic Communications}
\vspace{-1em}
\name{Ke~Yang $^{\star}$, Sixian~Wang$^{\star}$, Jincheng~Dai$^{\star}$, Kailin~Tan$^{\star}$, Kai~Niu$^{\star \dagger}$, Ping~Zhang$^{\star}$
\vspace{-0.4cm}
\thanks{This work was supported in part by the National Natural Science Foundation of China under Grant 92067202, Grant 62001049, Grant 62071058, and Grant 61971062, Beijing Natural Science Foundation under Grant 4222012, and the Fundamental Research Funds for the Central Universities.}
}

\address{
	\normalsize $^{\star}$ Beijing University of Posts and Telecommunications, Beijing, China \\
	\normalsize $^{\dagger}$ Peng Cheng Laboratory, Shenzhen, China\\
 \normalsize Email: daijincheng@bupt.edu.cn
}
\begin{document}
\ninept
\maketitle

\begin{abstract}
In this paper, we aim to redesign the vision Transformer (ViT) as a new backbone to realize semantic image transmission, termed wireless image transmission transformer (WITT). Previous works build upon convolutional neural networks (CNNs), which are inefficient in capturing global dependencies, resulting in degraded end-to-end transmission performance especially for high-resolution images. To tackle this, the proposed WITT employs Swin Transformers as a more capable backbone to extract long-range information. Different from ViTs in image classification tasks, WITT is highly optimized for image transmission while considering the effect of the wireless channel. Specifically, we propose a spatial modulation module to scale the latent representations according to channel state information, which enhances the ability of a single model to deal with various channel conditions. As a result, extensive experiments verify that our WITT attains better performance for different image resolutions,  distortion metrics, and channel conditions. The code is available at https://github.com/KeYang8/WITT.
\end{abstract}


\section{Introduction}

Recently, learning-based joint source-channel coding (JSCC) for wireless data transmission emerges as an active research area in communication community \cite{choi2019neural,farsad2018deep,DJSCC,ADJSCC,xie2021deep,Dai9852388, zhang2022toward}.
By replacing the hand-crafted codecs with deep neural networks (DNNs), they achieve comparable or even better end-to-end transmission performance than traditional separation-based source and channel coding schemes. In particular, for image transmission tasks, deep JSCC \cite{DJSCC} and its variants \cite{DJSCCF, ADJSCC, junperceptual, jankowski2020wireless} have competitive performance and much lower complexity compared to advanced image codec (JPEG/JPEG2000/BPG) followed by capacity-approaching channel code family (such as low-density parity-check (LDPC) coding \cite{LDPC_5G}). Moreover, they can be agilely optimized for human visual perception \cite{junperceptual}, or downstream machine tasks \cite{jankowski2020wireless}. Therefore, it is promising for many latency-sensitive applications, such as XR and autonomous driving.

Despite its great potential, previous works mainly build upon CNNs \cite{DJSCC, DJSCCF, ADJSCC}. Limited by the model capacity, it can be observed that with the increase of the image dimension, the performance of the CNN-based deep JSCC degrades rapidly and falls behind separation-based schemes. In this paper, we aim to break the aforementioned limits and increase the representation capacity of deep JSCC models.
To this end, we expect to introduce the global attention mechanism among all the image patches to extract \emph{high-level semantic features} of the source image for boosting a more efficient wireless image transmission method. Inspired by the recent advances of vision Transformer \cite{dosovitskiy2020image} in the computer vision field, it is the very time to redesign the vision Transformer as a new backbone for wireless image transmission.

In this paper, we propose a new JSCC framework named WITT, a high-efficiency wireless image transmission scheme that injects the advantages of the vision Transformer into the deep JSCC framework. By integrating the Swin Transformer \cite{liu2021swin} backbone in our scheme, a considerable performance gain can be achieved, especially for high-resolution images. Swin Transformer constructs hierarchical feature maps in the latent semantic space and has linear computational complexity to image size. Nevertheless, a naive change of the network backbone cannot obtain the expectable transmission performance gain over the imperfect wireless channels. To tackle this, we design a plug-in ``Channel ModNet'' inserted into Transformer to track the varying channel states. By this means, a single model can adapt to various channel states without retraining, which makes sense for the practical use of WITT.

We verify the performance of the proposed method through extensive experiments. We show that for image transmission, the proposed WITT method can achieve significant performance on various metrics such as PSNR and MS-SSIM \cite{ding2021comparison}. Equivalently, the proposed method can save bandwidth costs by achieving identical end-to-end transmission performance. As the source image resolution increases, the performance superiority shows more clearly.

\begin{figure*}[ht]
\vspace{-2em}
	\setlength{\abovecaptionskip}{0.cm}
	\setlength{\belowcaptionskip}{-0.cm}
	\centering{\includegraphics[scale=0.265]{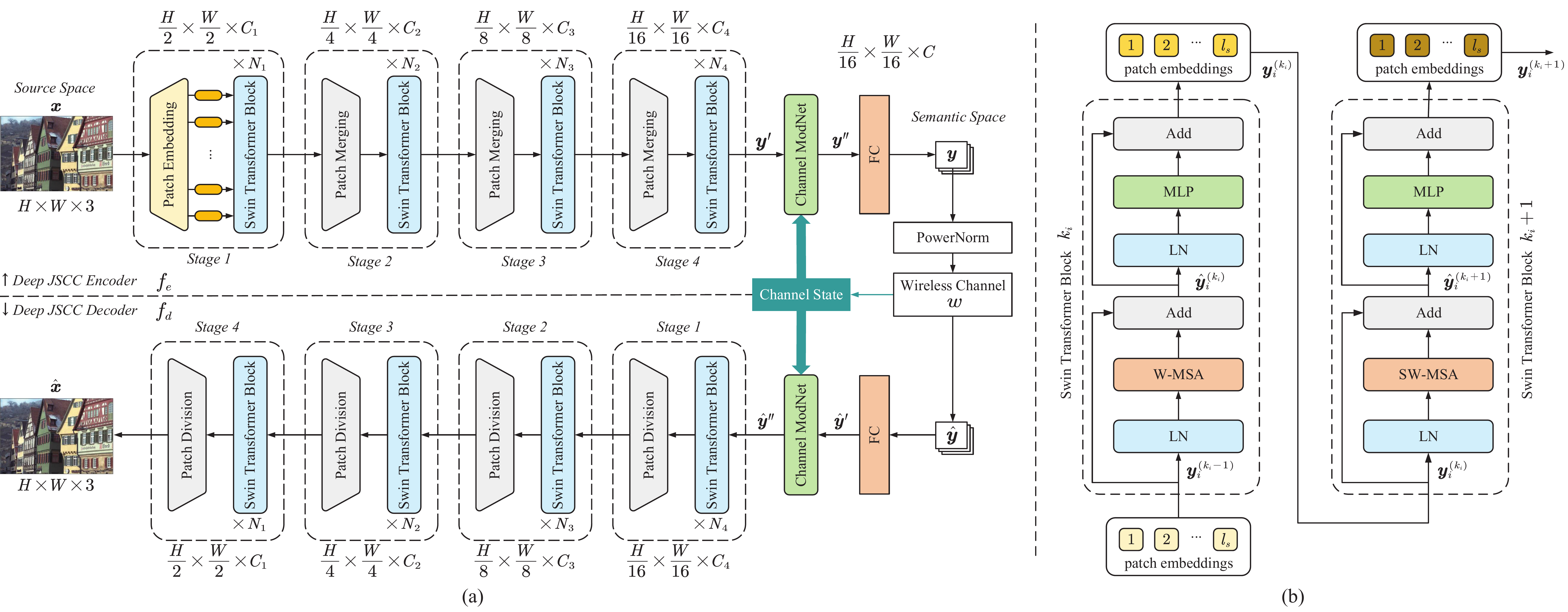}}
 \vspace{-1em}
	\caption{(a) The overall architecture of the proposed WITT scheme for wireless image transmission. (b) Two successive Swin Transformer Blocks. W-MSA and SW-MAS are multi-head self attention modules with regular and shifted windowing configurations, respectively.}\label{Fig1}
	\vspace{-1em}
\end{figure*}

\section{THE PROPOSED WITT SCHEME}\label{section_proposed_method}

\subsection{Overall Architecture}

An overview of the WITT architecture for wireless image transmission is given in Fig. \ref{Fig1}(a). An RGB image $\boldsymbol{x} \in \mathbb{R}^{H\times W\times3}$ is first split into $l_1=\frac{H}{2} \times \frac{W}{2}$ non-overlapping patches. Each patch can be viewed as a ``token''. We thus have a sequence of tokens $( x_1, \dots, x_{l_1} )$ by putting these tokens in the order from top left to bottom right. After patch embedding, $N_1$ Swin Transformer blocks are applied on these $l_1$ tokens \cite{liu2021swin}. Here, we refer to these $N_1$ blocks together with the patch embedding layer as ``stage 1''.

Then, these tokens are fed to several stages where ``stage $i$'' is encapsulated by a down-sampling patch merging layer and the following $N_i$ Swin Transformer blocks. It is worth noting that the number of stages of the encoder $f_e$ should vary with the image resolution. In general, images of higher resolution need more stages. As an example, Fig. \ref{Fig1}(a) presents the four-stages version. After that, each patch embedding will be rescaled by a Channel ModNet according to the channel state. Next, an FC layer is applied on these embeddings to project it to $C$ dimension. The channel bandwidth ratio is defined as $R= C / (2 \times 3 \times 2^n \times 2^n)$, where $n$ denotes the number of stage.

Before transmitting $\boldsymbol{y}$ into the wireless channel, the power normalization operation enables $\boldsymbol{y}$ to satisfy the average power constraint. Then, the analog feature map is directly sent over the wireless channel. In this paper, we consider the general fading channel model with transfer function $\boldsymbol{\hat{y}} = w( \boldsymbol{y} ; \boldsymbol{h} ) = \boldsymbol{h} \odot \boldsymbol{y} + \boldsymbol{n}$, where $\odot$ is the element-wise product, $\boldsymbol{h}$ denotes the channel state information (CSI) vector, and each component of the noise vector $\boldsymbol{n}$ is independently sampled from a Gaussian distribution, i.e., $\boldsymbol{n} \sim \mathcal{N}(0, {\sigma_n^2}{\boldsymbol{I}}_k)$, where ${\sigma_n^2}$ is the average noise power.

The deep JSCC decoder $f_d$ has a symmetric architecture with encoder $f_e$. It consists of an FC layer, Channel ModNet, patch division layers for up-sampling, and Swin Transformer blocks. It reconstructs input images from noisy latent representations $\boldsymbol{\hat{y}}$. The training loss function of the whole system is
\begin{equation}\label{eq_RD_function_target}
  \mathop{\min }\limits_{{\bm{\phi}},{\bm{\theta}}}{{\mathbb{E}}_{\boldsymbol{x} \sim {p_{\boldsymbol{x}}}}}{{\mathbb{E}}_{{\boldsymbol{\hat{y}}} \sim p_{{\boldsymbol{\hat{y}}} | {\boldsymbol{x}} }}}{\left[ {d\left( {{\boldsymbol{x}},{\boldsymbol{\hat{x}}}} \right)} \right]},
\end{equation}
where ${\boldsymbol{\hat y}} = w\left( f_{e}( {\boldsymbol x}; {\bm \phi} ) ; \bm{\nu} \right)$, ${\boldsymbol{\hat{x}}} = f_{d}(\boldsymbol{\hat y}; \bm{\theta})$, ${\bm{\phi}}$ and ${\bm{\theta}}$ encapsulate all the network parameters of $f_e$ and $f_d$, respectively.

\subsection{Swin Transformer Block}

As shown in Fig. \ref{Fig1}(b), a Swin Transformer block is a sequence-to-sequence function that is built by replacing the standard multi-head self attention (MSA) module in a Transformer block with a module based on shifted windows \cite{liu2021swin}. The shift of the window partition between consecutive self attention layers provides connections among them, significantly enhancing modeling power.

With the shifted window partitioning approach, consecutive Swin Transformer blocks $k_i$ and $k_i + 1$ of ``stage $i$'' are computed as
\begin{subequations}
\begin{equation}
    {\mathbf{\hat y}}_i^{({k_i})} = {\text{W-MSA}}({\text{LN}}({\mathbf{y}}_i^{({k_i} - 1)})) + {\mathbf{y}}_i^{({k_i} - 1)},
    \vspace{-0.8em}
\end{equation}
\begin{equation}
    {\mathbf{y}}_i^{({k_i})} = {\text{MLP}}({\text{LN}}({\mathbf{\hat y}}_i^{({k_i})})) + {\mathbf{\hat y}}_i^{({k_i})},
    \vspace{-0.3em}
\end{equation}
\end{subequations}
\begin{subequations}
\begin{equation}
    {\mathbf{\hat y}}_i^{({k_i}+1)} = {\text{SW-MSA}}({\text{LN}}({\mathbf{y}}_i^{({k_i}+1)})) + {\mathbf{y}}_i^{({k_i}+1)},
    \vspace{-0.3em}
\end{equation}
\begin{equation}
    {\mathbf{y}}_i^{({k_i}+1)} = {\text{MLP}}({\text{LN}}({\mathbf{\hat y}}_i^{({k_i}+1)})) + {\mathbf{\hat y}}_i^{({k_i}+1)},
    \vspace{-0.3em}
\end{equation}
\end{subequations}
where ${\mathbf{\hat y}}_i^{({k_i})}$ and ${\mathbf{y}}_i^{({k_i})}$ represent the output feature of the (S)W-MSA module and the MLP module for block $k_i$ at stage $i$, respectively. W-MSA and SW-MAS are multi-head self attention modules with regular and shifted windowing configurations. LN denotes the layer normalization operation \cite{dosovitskiy2020image}.

\subsection{Channel ModNet}

\begin{figure}[h]
	\setlength{\abovecaptionskip}{0.cm}
	\setlength{\belowcaptionskip}{-0.cm}
	\centering{\includegraphics[scale=0.54]{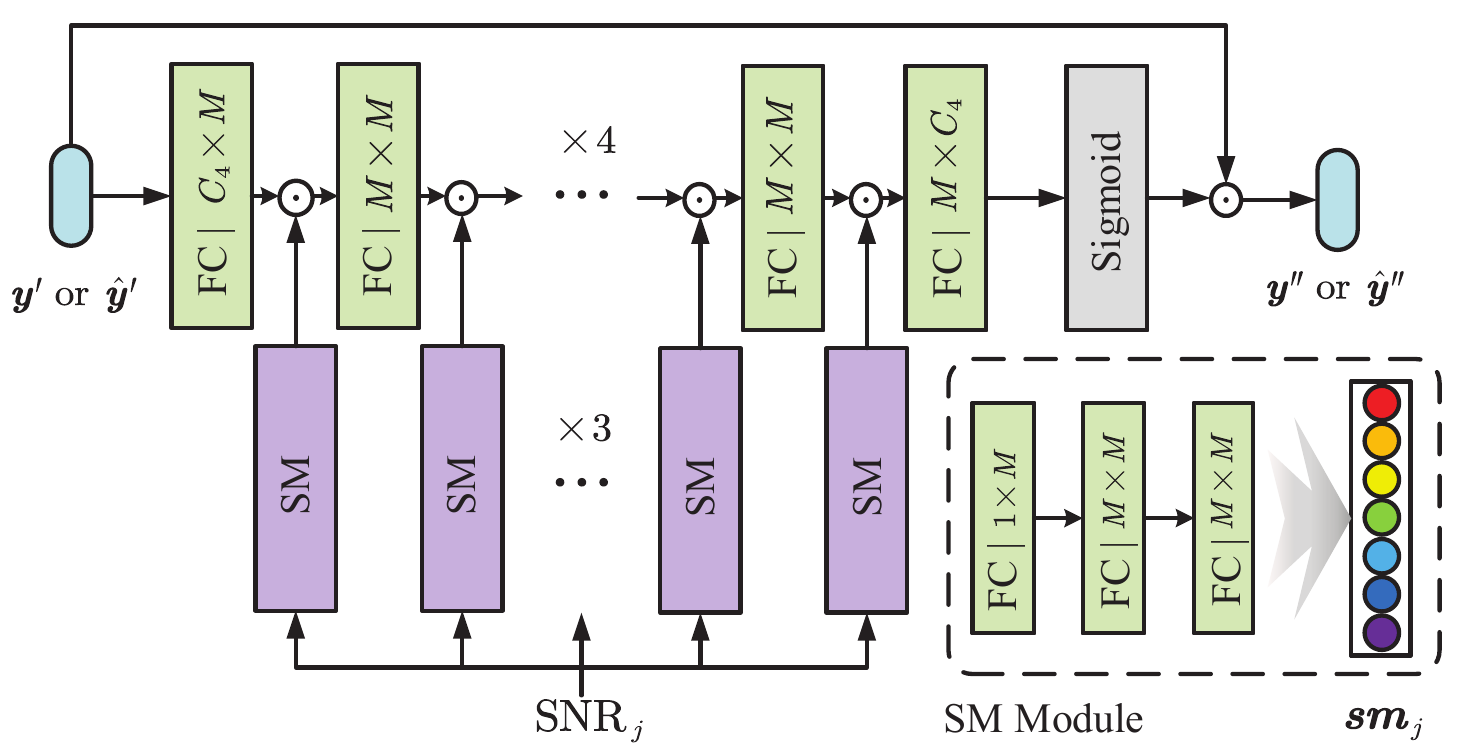}}
 \vspace{-1em}
	\caption{The architecture of Channel ModNet. $C_4$ and $M$ denote the number of channels in $\boldsymbol{y}'$ or $\boldsymbol{\hat{y}}'$ and the number of intermediates of FCN in $\boldsymbol{sm}_j$ respectively. } \label{Fig2}
	\vspace{-1em}
\end{figure}

The proposed ``Channel ModNet'' is a plug-in module to modulate the output of several Transformer stages as shown in Fig. \ref{Fig1}(a). For different channel states, Channel ModNet can generate specific deep JSCC codec functions to adapt to channel changes.

In particular, the semantic feature map $\boldsymbol{y}'$ is fed into our Channel ModNet to be modulated by the channel state information. Correspondingly, the received symbol $\boldsymbol{\hat{y}}$ is first processed by the FC layer and then sent into our ModNet. Thus, for both $f_e$ and $f_d$, the channel state is taken as a coding factor sent into Channel ModNet, which modulates the intermediate feature maps to the wireless channel state.

\begin{figure*}[htbp]
    \setlength{\abovecaptionskip}{0.cm}
    \setlength{\belowcaptionskip}{-0.cm}
\begin{center}
 \subfigure[]{
\includegraphics[width=0.166\linewidth]{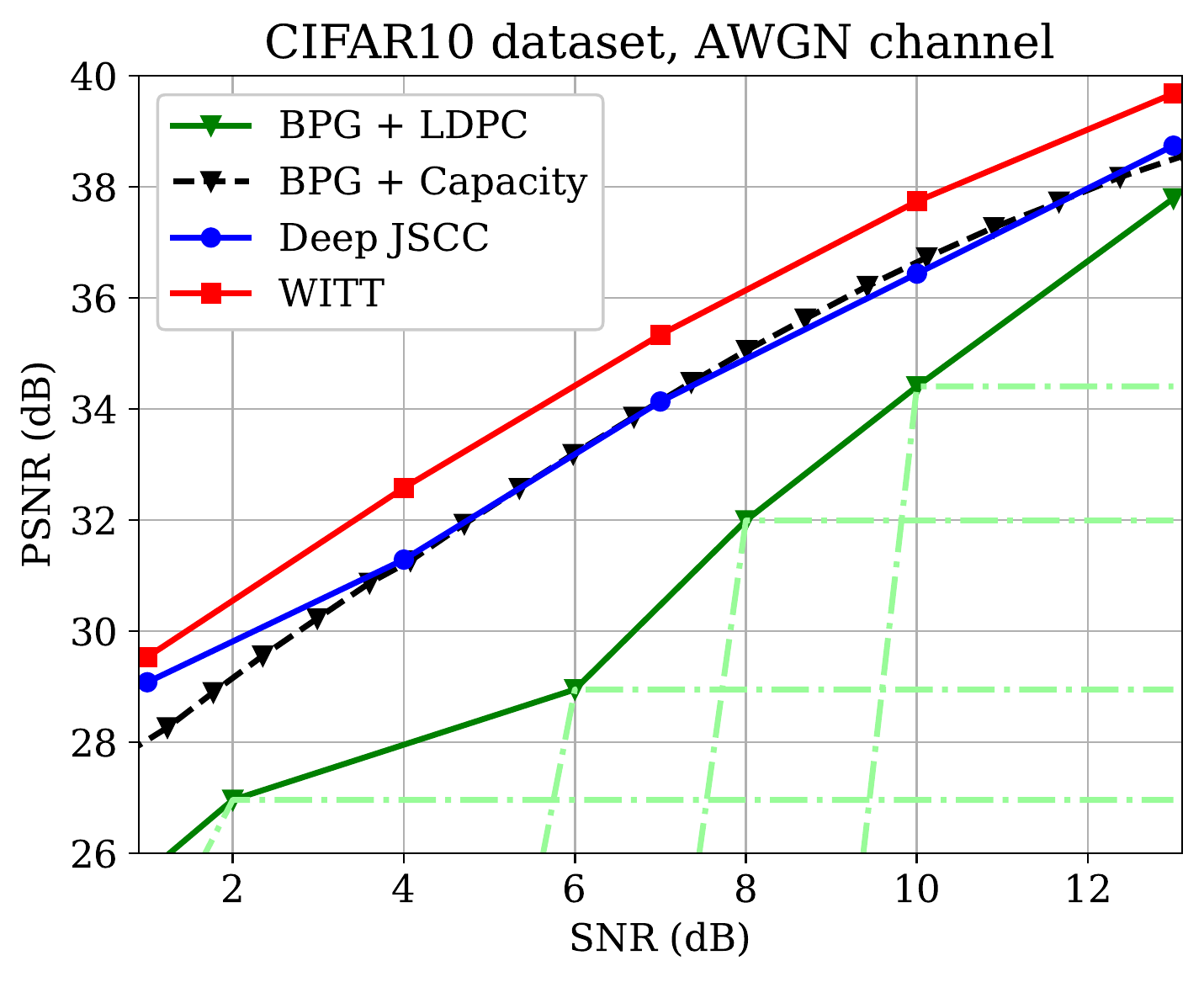}}
\hspace{-.1in}
\subfigure[]{
\includegraphics[width=0.166\linewidth]{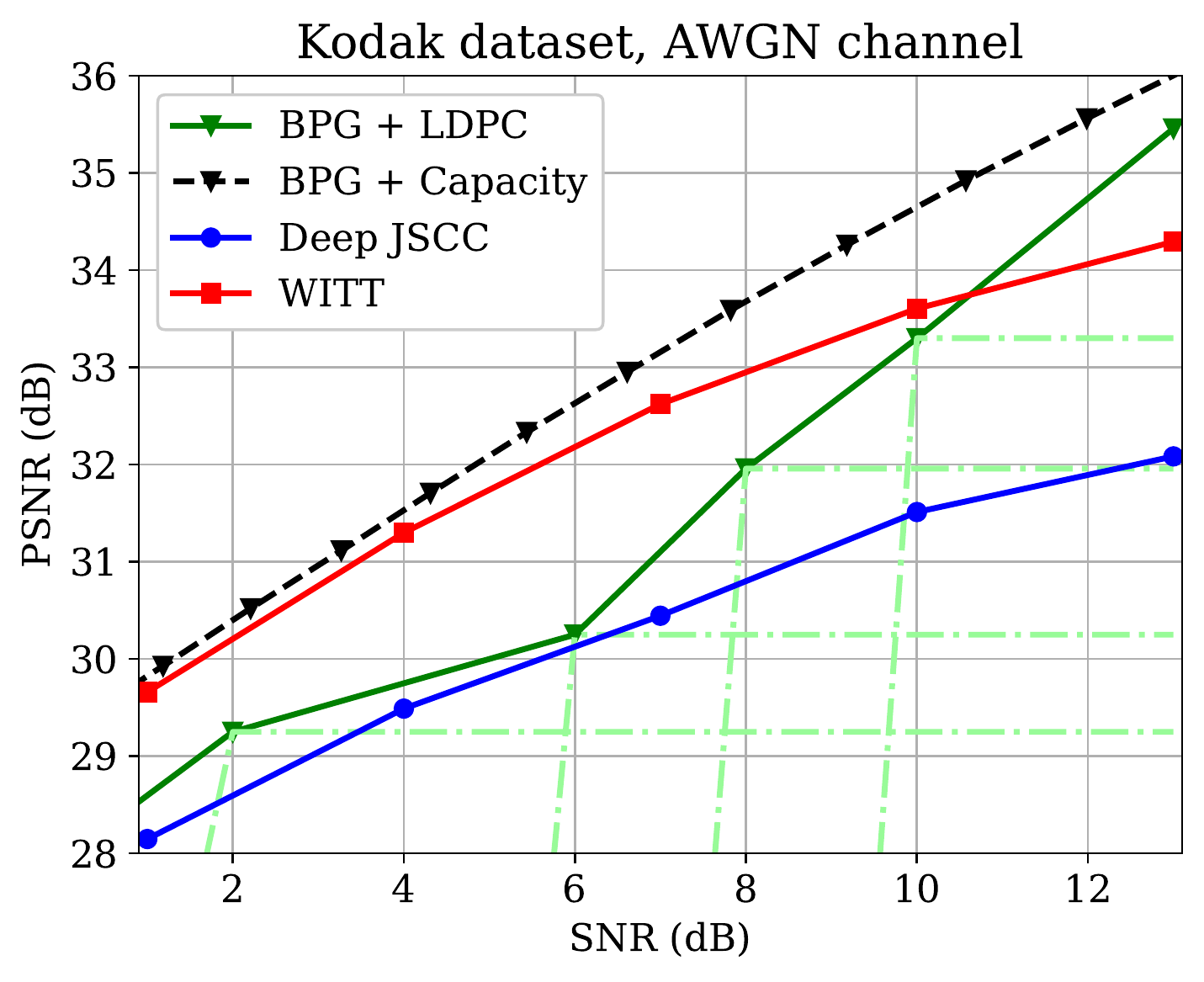}}
\hspace{-.10in}
\subfigure[]{
\includegraphics[width=0.166\linewidth]{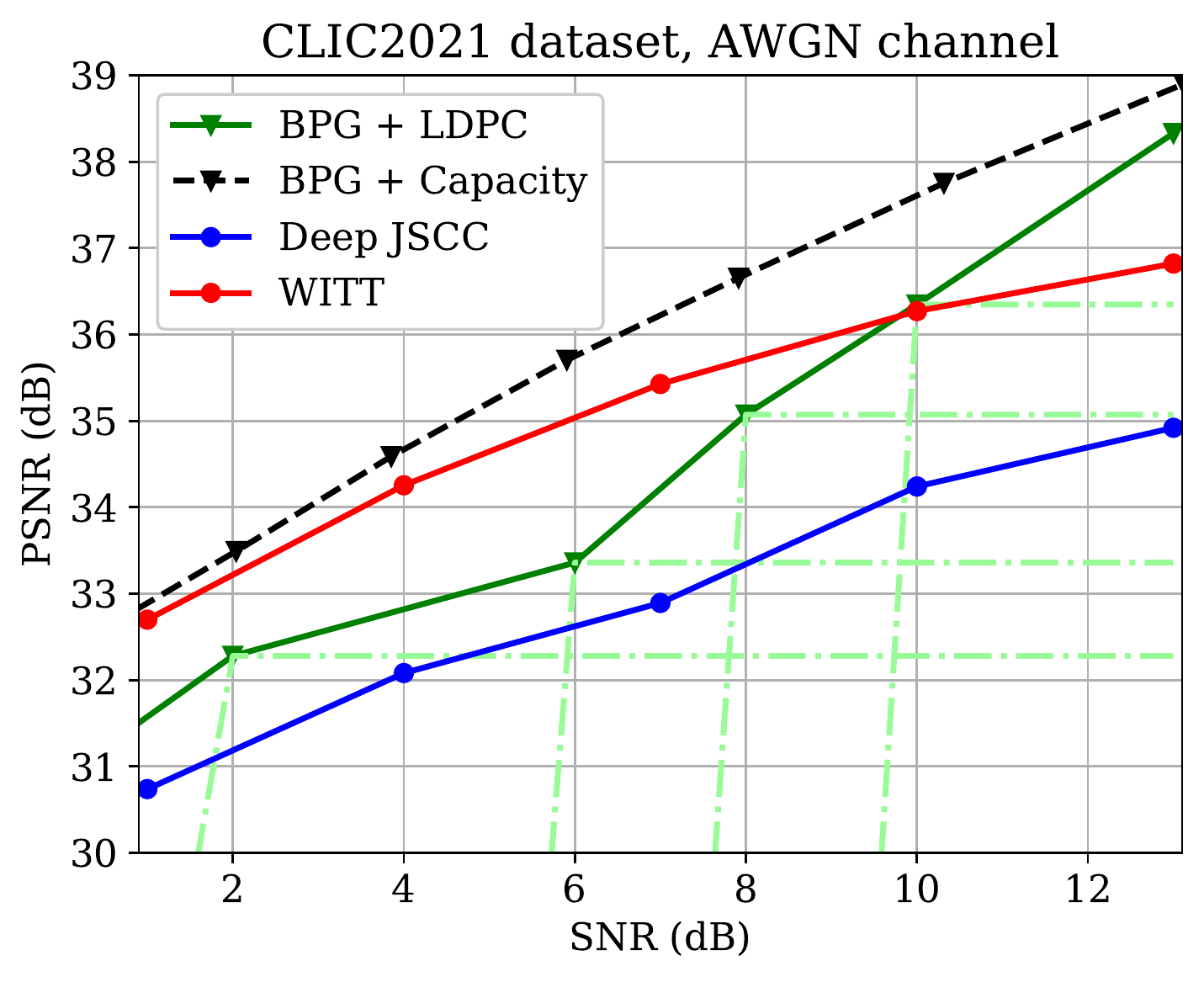}}
\hspace{-.10in}
 \subfigure[]{
\includegraphics[width=0.166\linewidth]{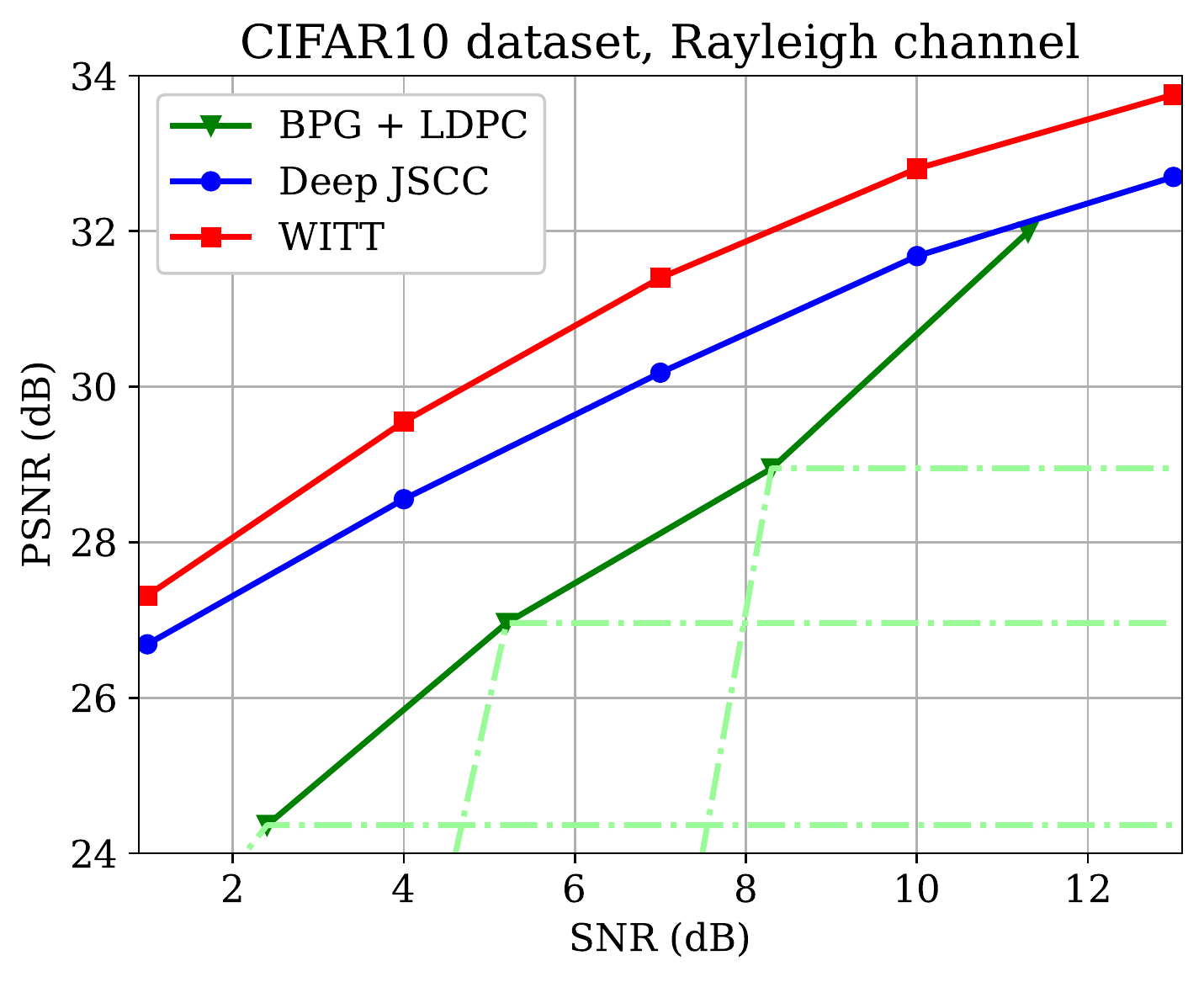}}
\hspace{-.10in}
\subfigure[]{
\includegraphics[width=0.166\linewidth]{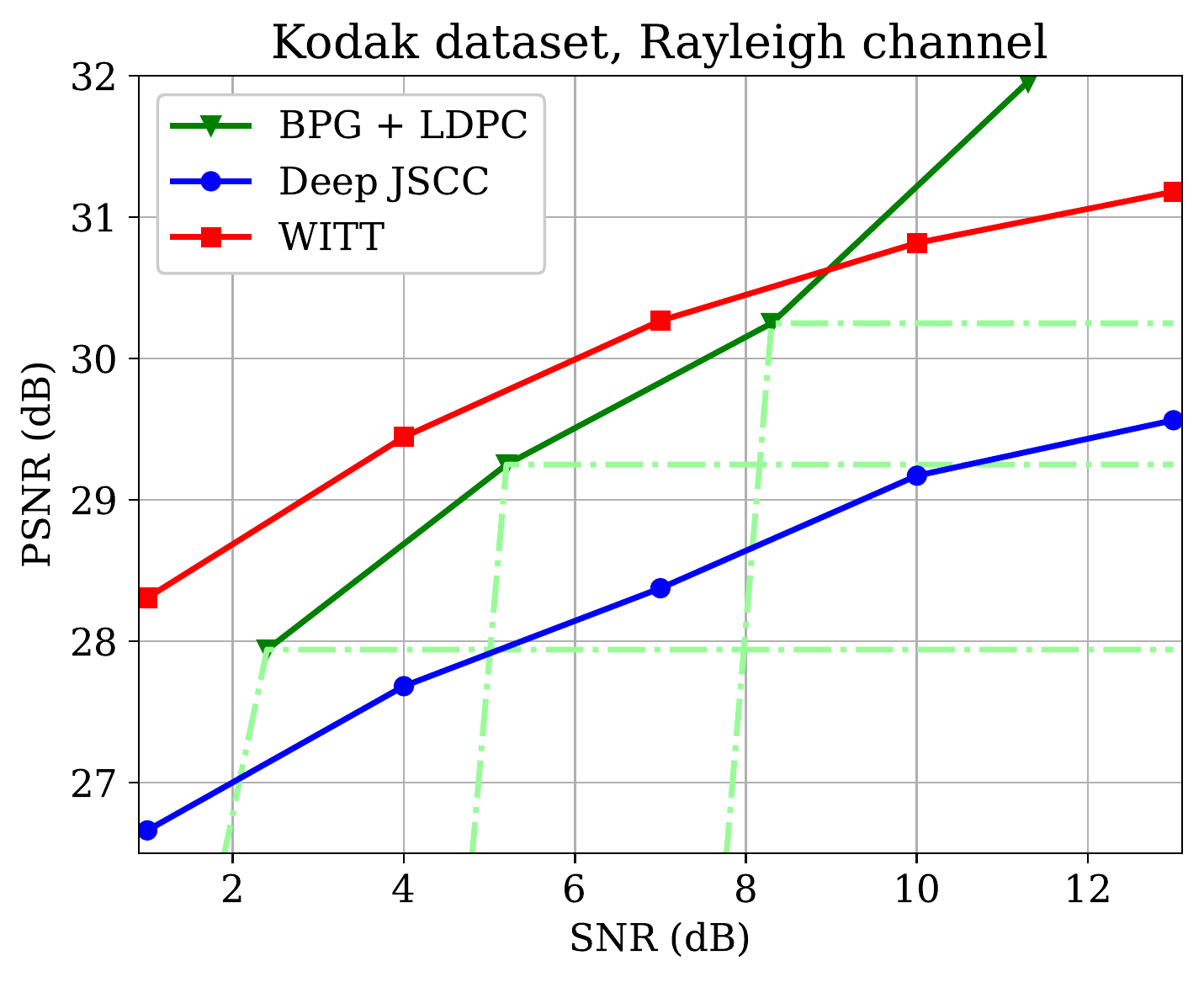}}
\hspace{-.10in}
\subfigure[]{
\includegraphics[width=0.166\linewidth]{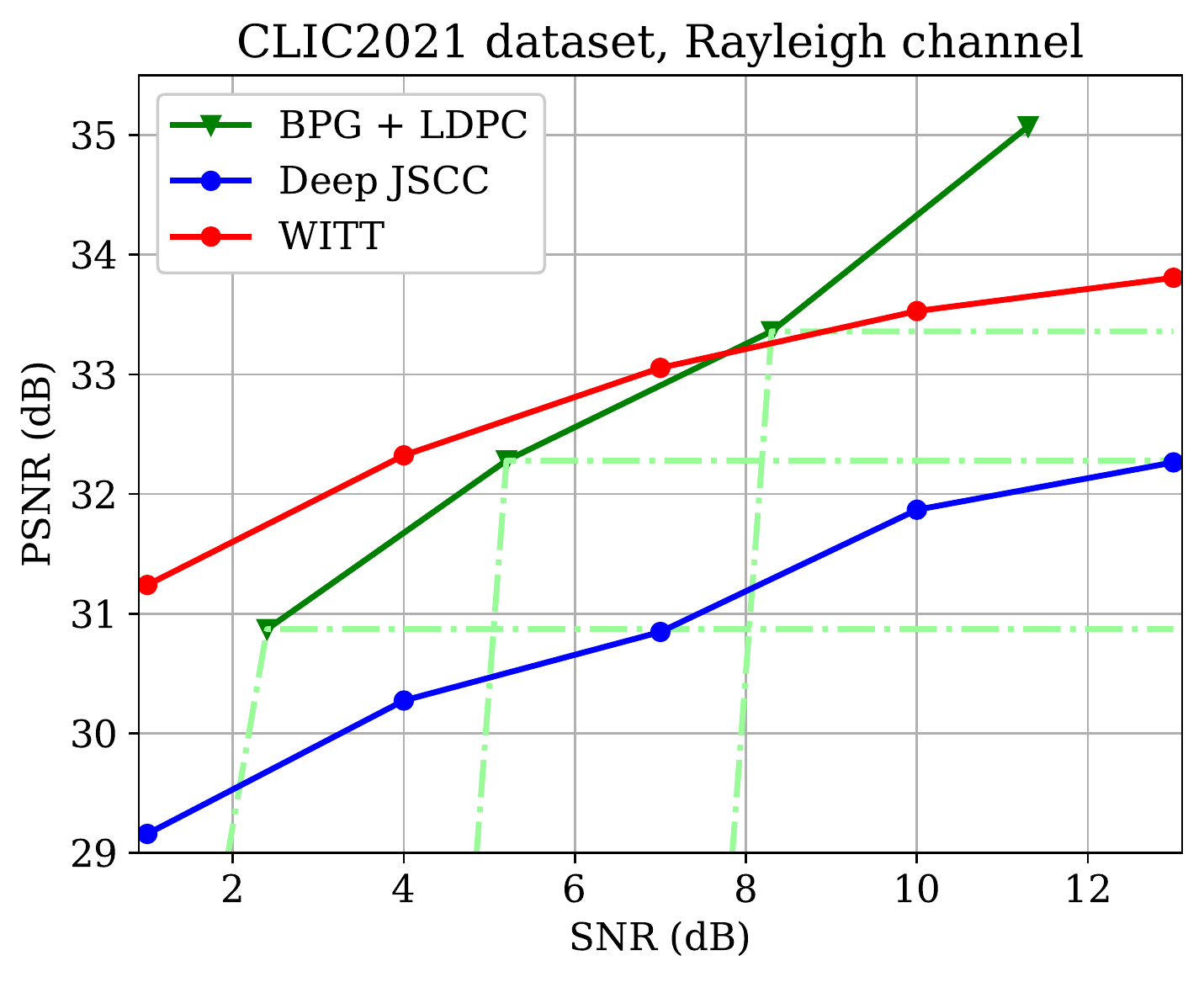}}
\hspace{-.10in}
\vspace{-1em}
\caption{(a)$\sim$(c) PSNR performance versus the SNR over the AWGN channel. (d)$\sim$(f) PSNR performance versus the SNR over the Rayleigh fast fading channel. The average CBR is set to $1/3$, $1/16$, and $1/16$ for CIFAR10 dataset, Kodak dataset, and CLIC2021 dataset.}\label{Fig3}
\vspace{-1.5em}
\end{center}
\end{figure*}

\begin{figure*}[htbp]
    \setlength{\abovecaptionskip}{0.cm}
    \setlength{\belowcaptionskip}{-0.cm}
\begin{center}
 \subfigure[]{
\includegraphics[width=0.166\linewidth]{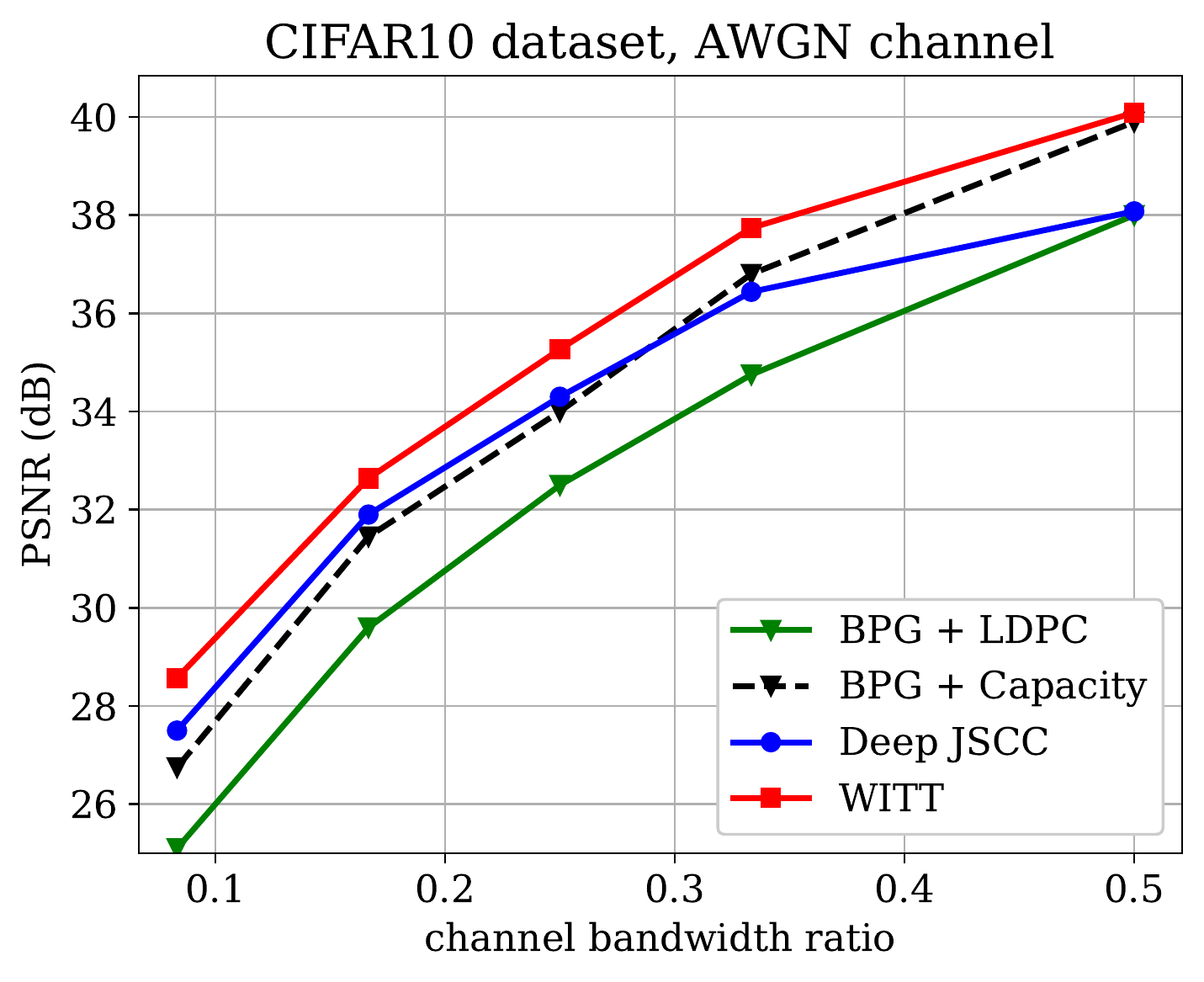}}
\hspace{-.1in}
\subfigure[]{
\includegraphics[width=0.166\linewidth]{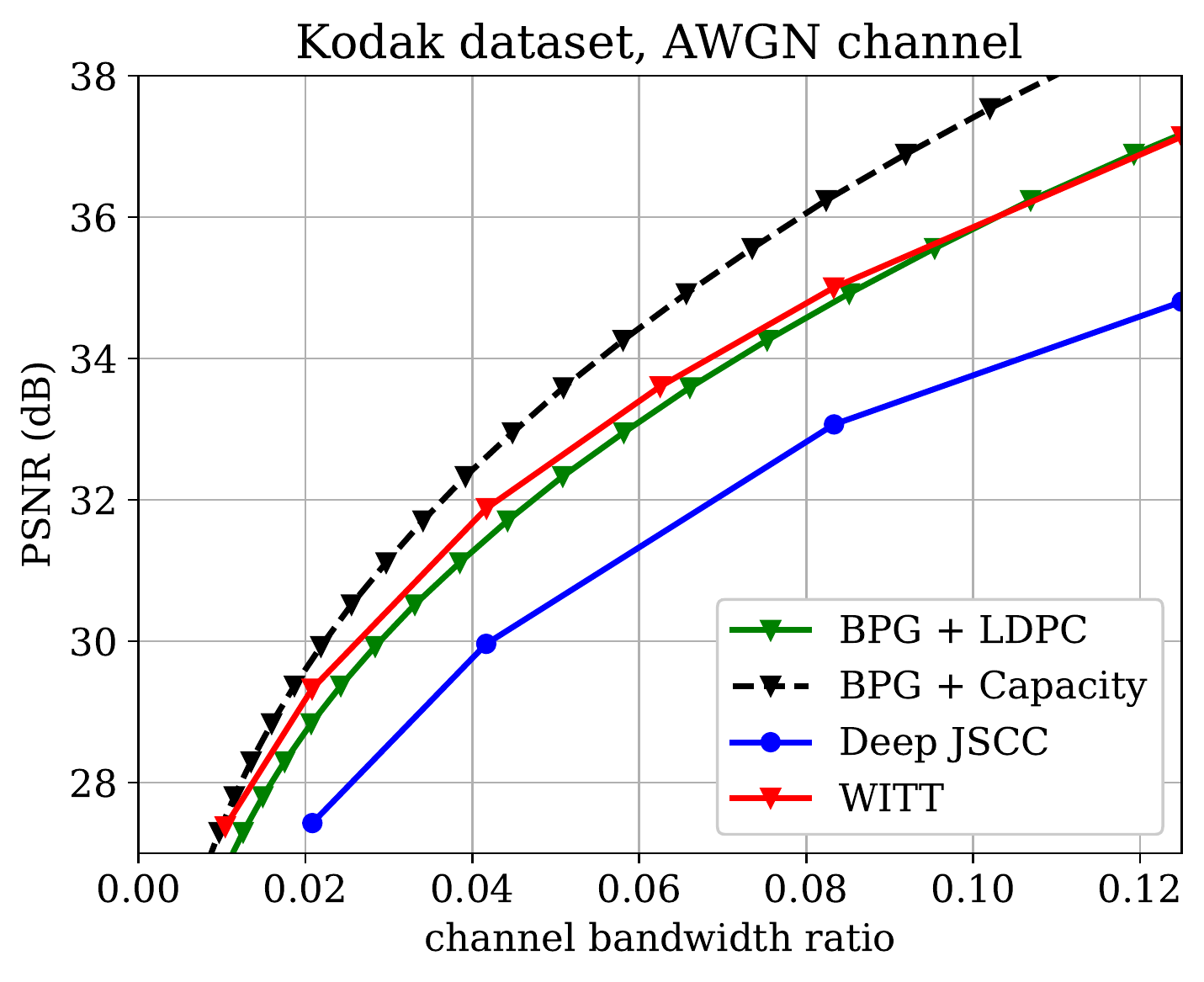}}
\hspace{-.10in}
\subfigure[]{
\includegraphics[width=0.166\linewidth]{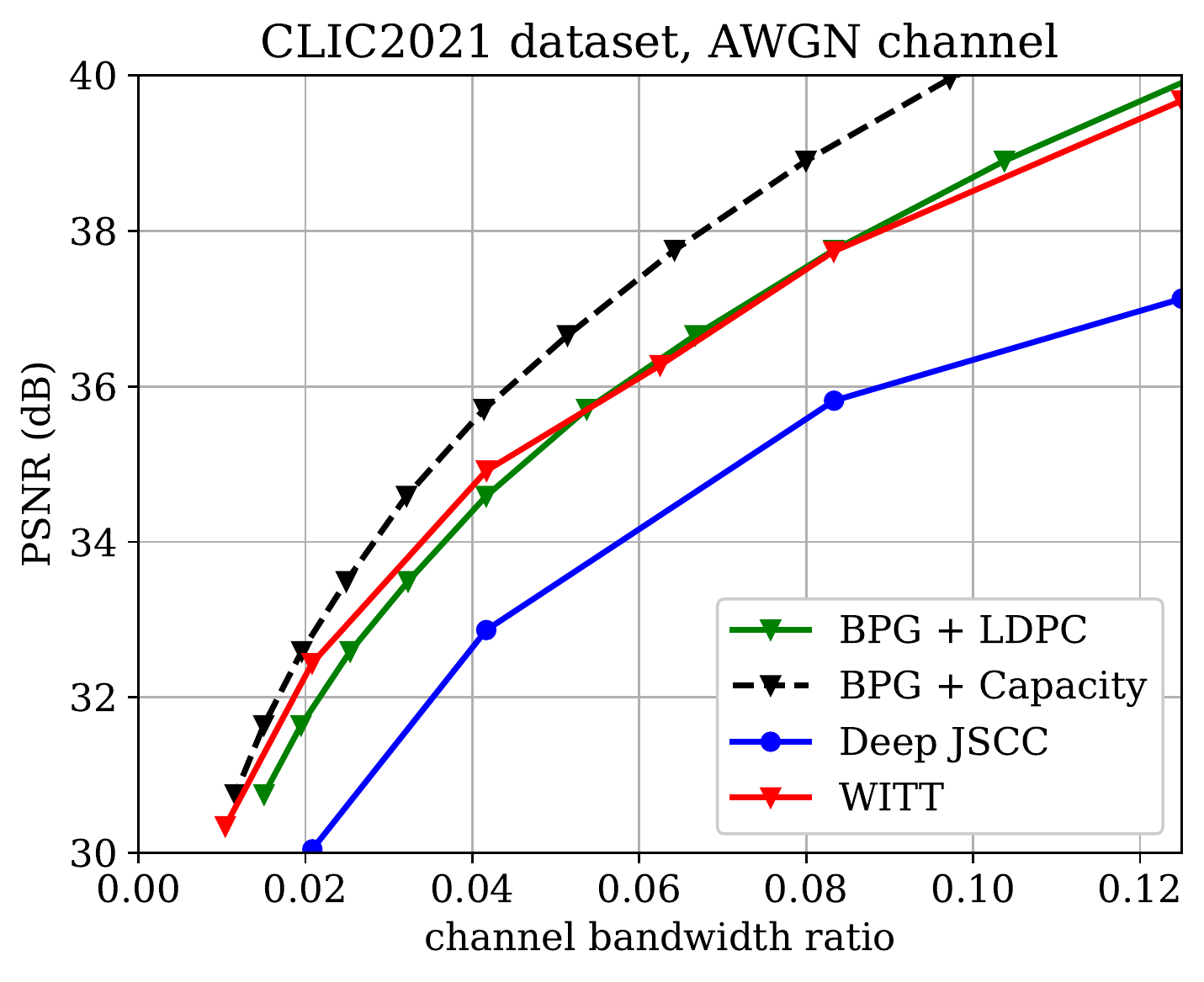}}
\hspace{-.10in}
 \subfigure[]{
\includegraphics[width=0.166\linewidth]{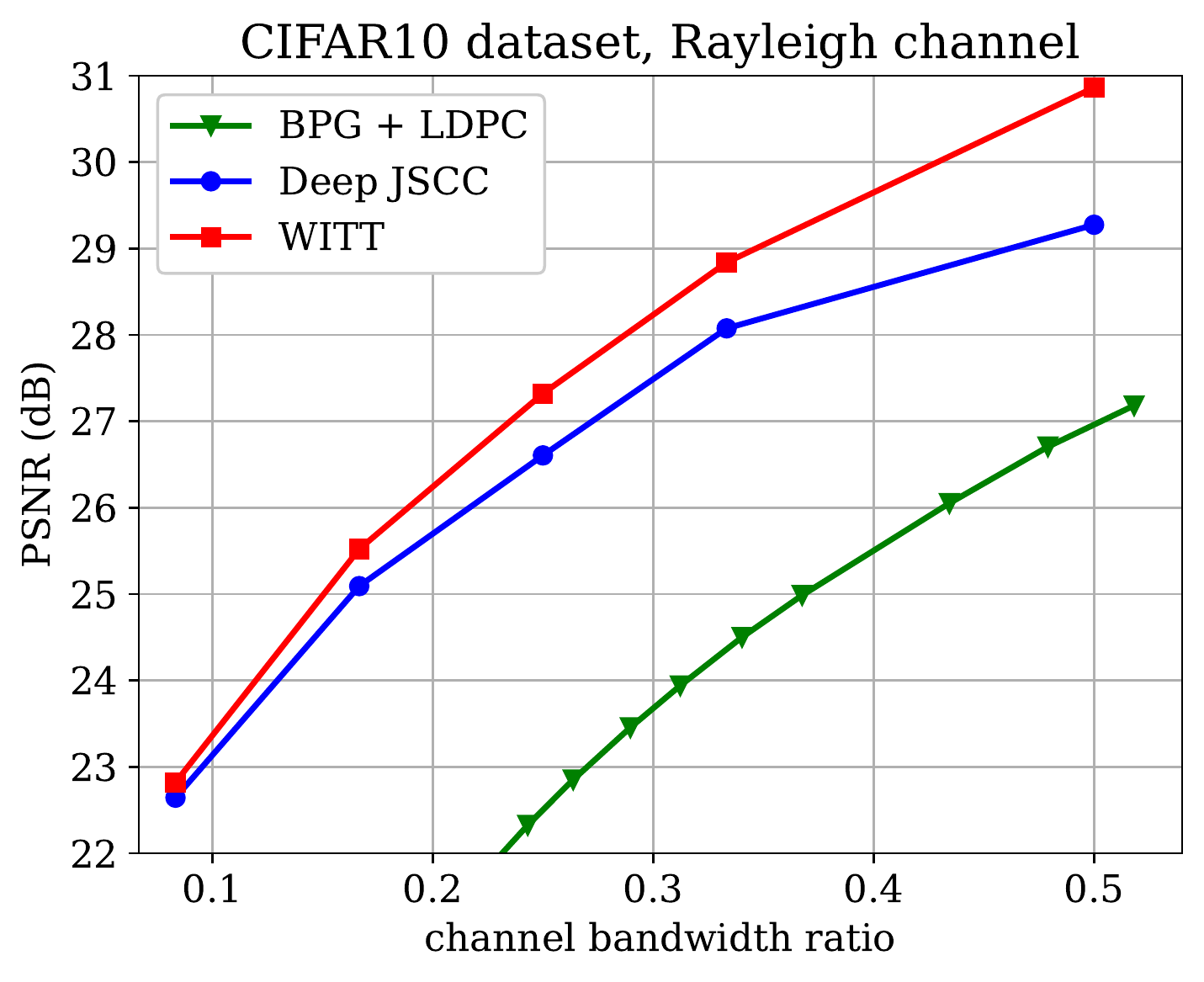}}
\hspace{-.10in}
\subfigure[]{
\includegraphics[width=0.166\linewidth]{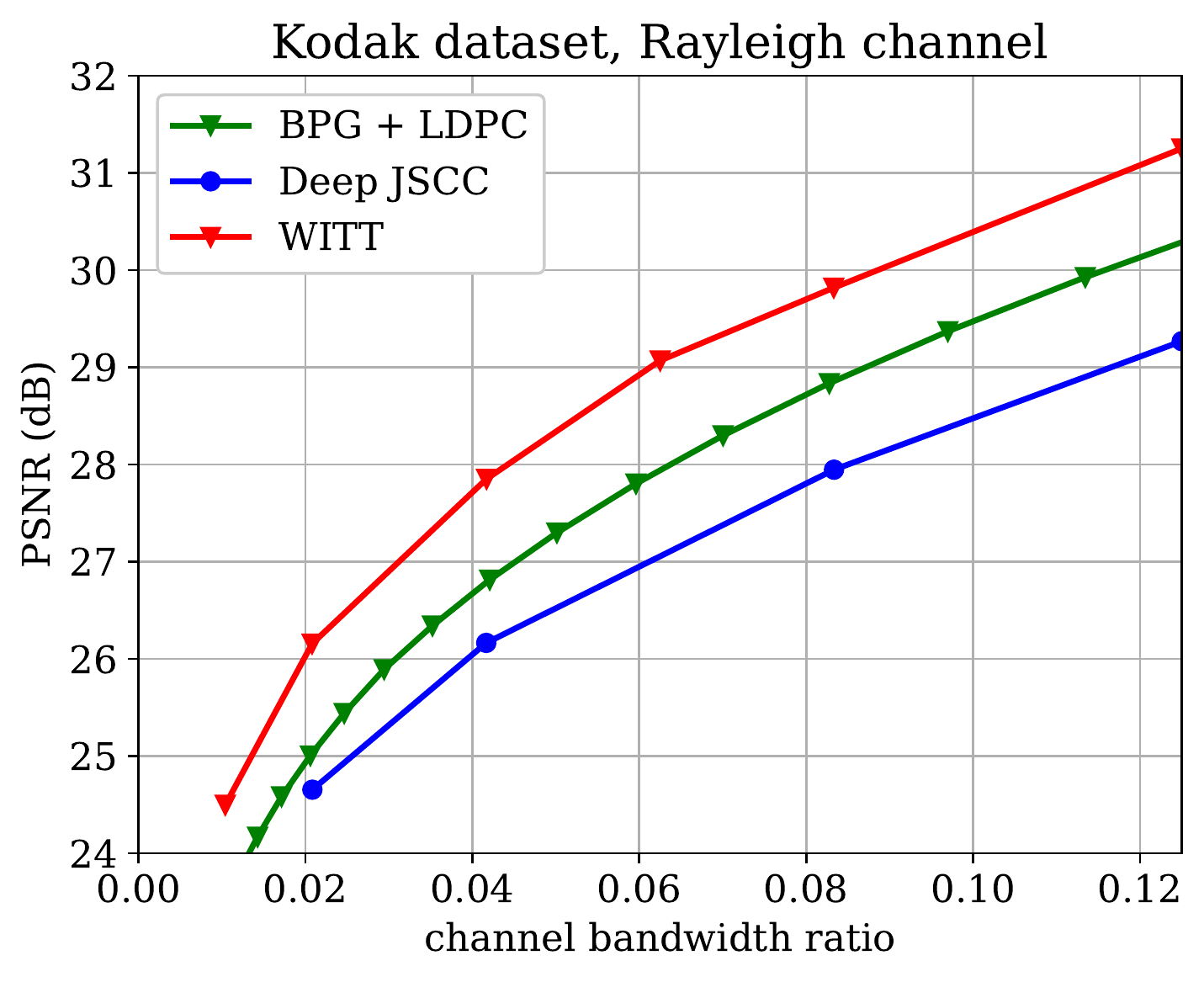}}
\hspace{-.10in}
\subfigure[]{
\includegraphics[width=0.166\linewidth]{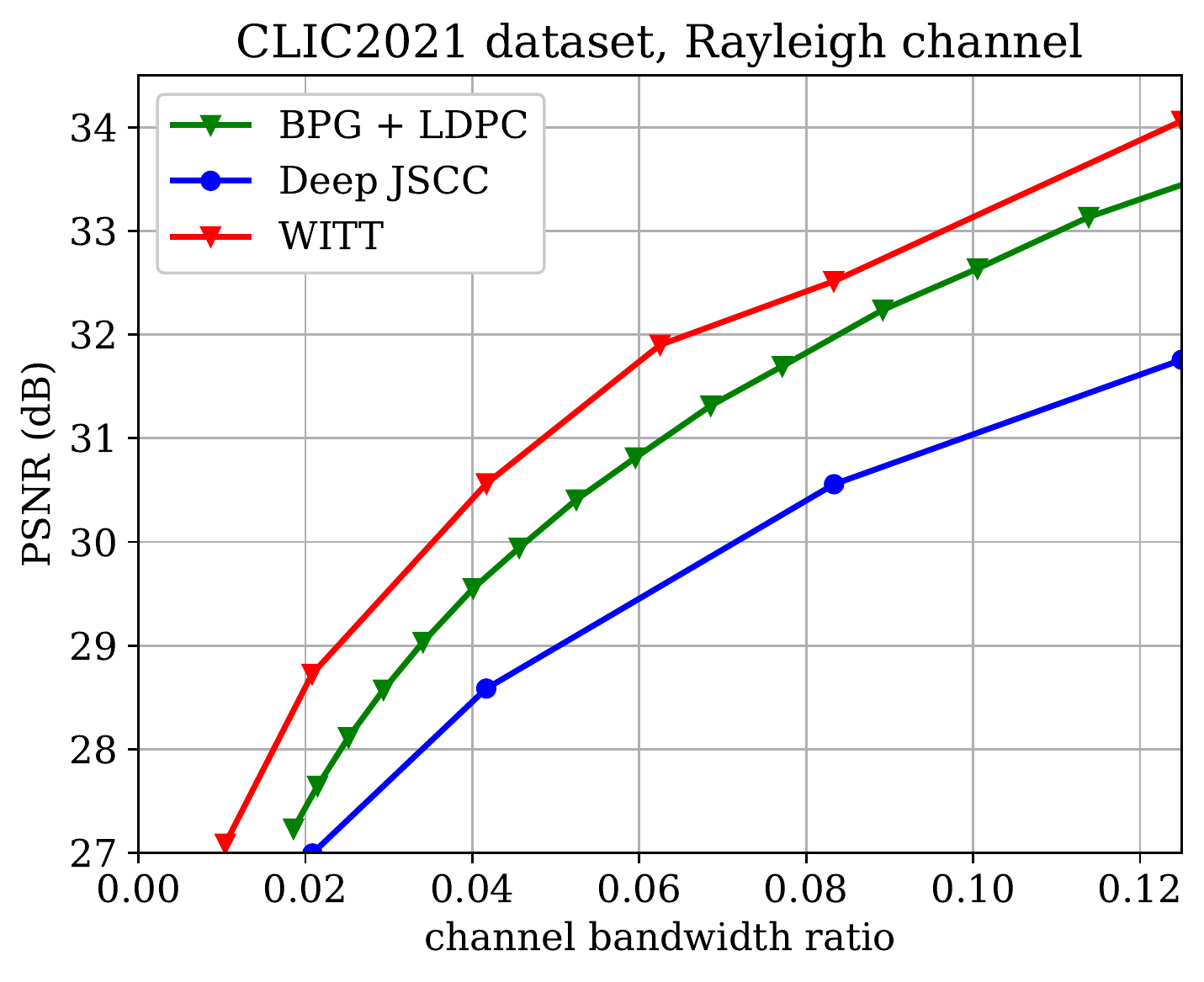}}
\hspace{-.10in}
\vspace{-1em}
\caption{(a)$\sim$(c) PSNR performance versus the CBR over the AWGN channel at $\text{SNR} = 10\text{dB}$. (d)$\sim$(f) PSNR performance versus the CBR over the Rayleigh fast fading channel at $\text{SNR} = 3\text{dB}$.}\label{Fig4}
\vspace{-2.5em}
\end{center}
\end{figure*}

The architecture of the Channel ModNet is depicted in Fig. \ref{Fig2}. It consists of 8 FC layers alternating with 7 SNR modulation (SM) modules. SM module is a three-layered FC network, which transforms the input $\text{SNR}_j$ into an $M$-dimensional vector $\boldsymbol{sm}_j$. Multiple SM modules are cascaded sequentially in a coarse-to-fine manner. The previous modulated features are fed into subsequent SM modules. The arbitrary target modulator can be realized by assigning a corresponding SNR value. The mapping procedures from $\text{SNR}_j$ to $\boldsymbol{sm}_j$ are
\vspace{-0.5em}
\begin{subequations}
    \begin{equation}
        \boldsymbol{sm}^{(1)}_j = \text{ReLU}(\boldsymbol{W}^{(1)} \cdot \text{SNR}_j + \boldsymbol{b}^{(1)}),
        \vspace{-0.5em}
    \end{equation}
    \begin{equation}
        \boldsymbol{sm}^{(2)}_j = \text{ReLU}(\boldsymbol{W}^{(2)} \cdot \boldsymbol{sm}^{(1)}_j + \boldsymbol{b}^{(2)}),
    \end{equation}
    \begin{equation}
        \boldsymbol{sm}_j = \text{ReLU}(\boldsymbol{W}^{(3)} \cdot \boldsymbol{sm}^{(2)}_j + \boldsymbol{b}^{(3)}),
    \end{equation}
\end{subequations}
where ReLU and Sigmoid are the activation functions, $\boldsymbol{W}$ and $\boldsymbol{b}$ are the affine function parameters and their corresponding bias.

Therefore, the coding $\text{SNR}_j$ is associated with a tensor $\boldsymbol{sm}_j$ in each SM module. Then, the input feature will be fused with $\boldsymbol{sm}_j$ in the element-wise product, i.e.,
\vspace{-0.5em}
\begin{equation}
      \boldsymbol{output} = \boldsymbol{input} \odot \boldsymbol{sm}_j
\vspace{-0.5em}
\end{equation}
Here, $\boldsymbol{input}$ denotes the feature output from the previous FC layer, and $\boldsymbol{output}$ is feeding into the next FC layer.

\makeatletter
\renewcommand{\@thesubfigure}{\hskip\subfiglabelskip}
\makeatother

\begin{figure*}[t]
	\setlength{\abovecaptionskip}{0cm}
	\setlength{\belowcaptionskip}{0cm}
	\begin{subtable}
		\centering
		\small
		\begin{tabular}{m{0.18\textwidth}m{0.12\textwidth}<{\centering}m{0.132\textwidth}<{\centering}m{0.1357\textwidth}<{\centering}m{0.1307\textwidth}<{\centering}m{0.1307\textwidth}<{\centering}}
			& Original & Deep JSCC & BPG + LDPC  &  BPG + Capacity & Our WITT  \\
		\end{tabular}
	\end{subtable}

	\begin{center}
 \vspace{-1em}

   \hspace{-.05in}
	\subfigure[$512\times768$] {\includegraphics[width=0.15\textwidth]{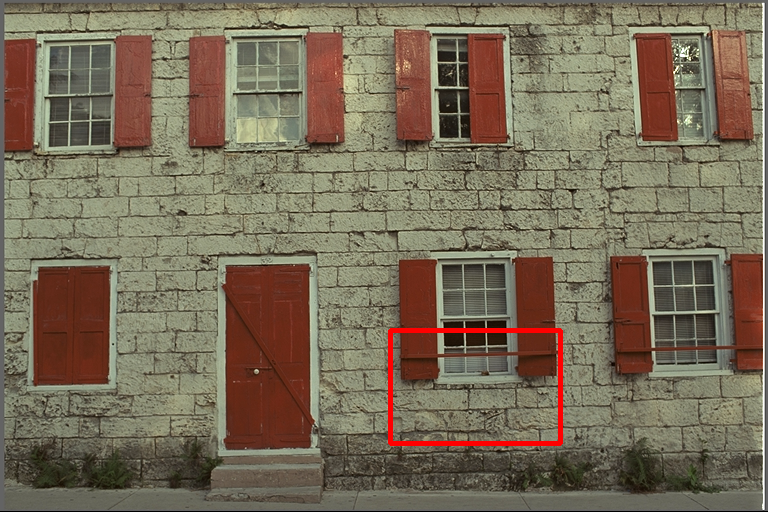}}
	\quad
	\subfigure[$R$ / PSNR (dB)] {\includegraphics[width=0.15\textwidth]{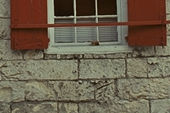}}
	\hspace{-.15in}
	\quad
	\subfigure[0.0208 (\textit{0\%}) / 24.81]{\includegraphics[width=0.15\textwidth]{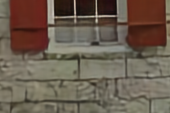}}
	\hspace{-.15in}
	\quad
	\subfigure[0.0237 (\textcolor{red}{\textit{+14\%}}) / 25.22] {\includegraphics[width=0.15\textwidth]{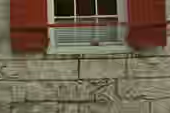}}
	\hspace{-.15in}
	\quad
	\subfigure[0.0217 (\textcolor{red}{\textit{+4\%}}) / 25.70] {\includegraphics[width=0.15\textwidth]{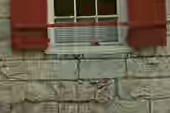}}
 	\hspace{-.15in}
	\quad
 \subfigure[0.0208 (\textit{0\%}) / 26.16] {\includegraphics[width=0.15\textwidth]{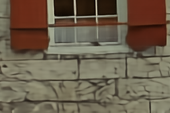}}

	   \hspace{-.05in}
	\subfigure[$512\times768$] {\includegraphics[width=0.15\textwidth]{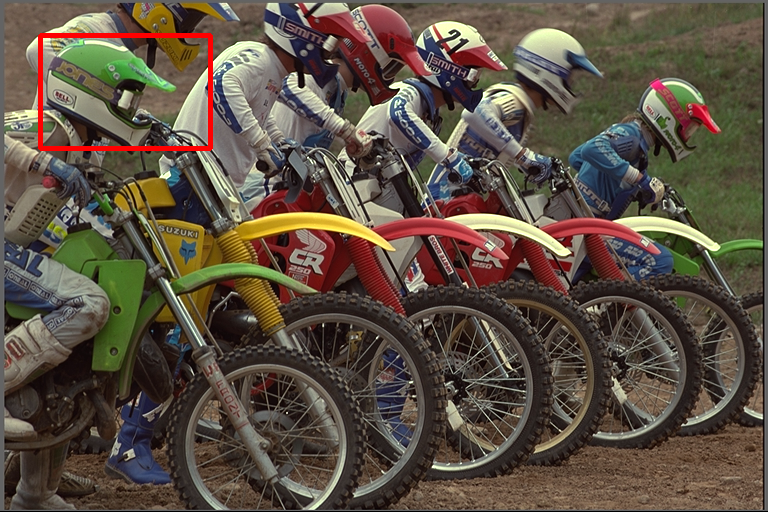}}
	\quad
	\subfigure[$R$ / PSNR (dB)] {\includegraphics[width=0.15\textwidth]{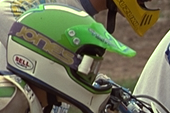}}
	\hspace{-.15in}
	\quad
	\subfigure[0.0208 (\textit{0\%}) / 23.76]{\includegraphics[width=0.15\textwidth]{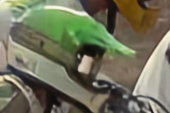}}
	\hspace{-.15in}
	\quad
	\subfigure[0.0209 (\textit{0\%}) / 23.37] {\includegraphics[width=0.15\textwidth]{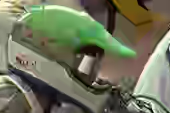}}
	\hspace{-.15in}
	\quad
	\subfigure[0.0223 (\textcolor{red}{\textit{+7\%}}) / 24.49] {\includegraphics[width=0.15\textwidth]{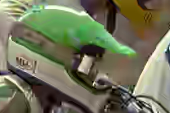}}
 	\hspace{-.15in}
	\quad
 \subfigure[0.0208 (\textit{0\%}) / 25.11] {\includegraphics[width=0.15\textwidth]{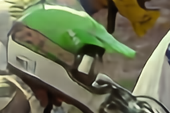}}
\vspace{-0.9em}
	\caption{Examples of visual comparison under AWGN channel at $\text{SNR} = 10$dB. The first column, second column, and third to sixth column shows the original image, original patch, and reconstructions of different transmission schemes, respectively. The red number indicates the percentage of extra bandwidth cost compared to WITT.}\label{Fig6}
	\vspace{-2em}
    \end{center}
\end{figure*}

\begin{figure}[tbp]
    \setlength{\abovecaptionskip}{0.cm}
    \setlength{\belowcaptionskip}{-0.cm}
\begin{center}
\hspace{-.10in}
\subfigure[]{
\includegraphics[width=0.5\linewidth]{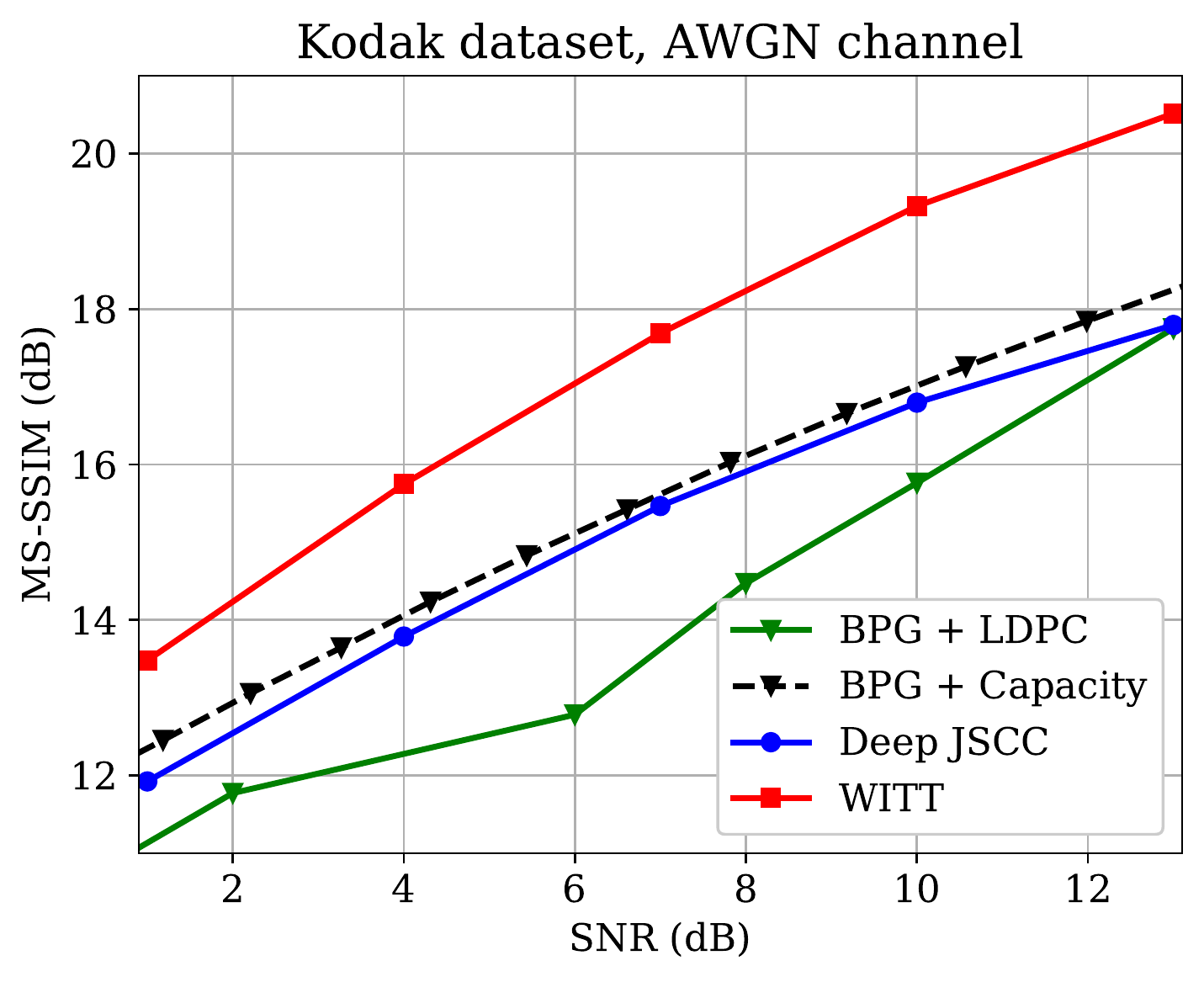}}
\hspace{-.10in}
\subfigure[]{
\includegraphics[width=0.5\linewidth]{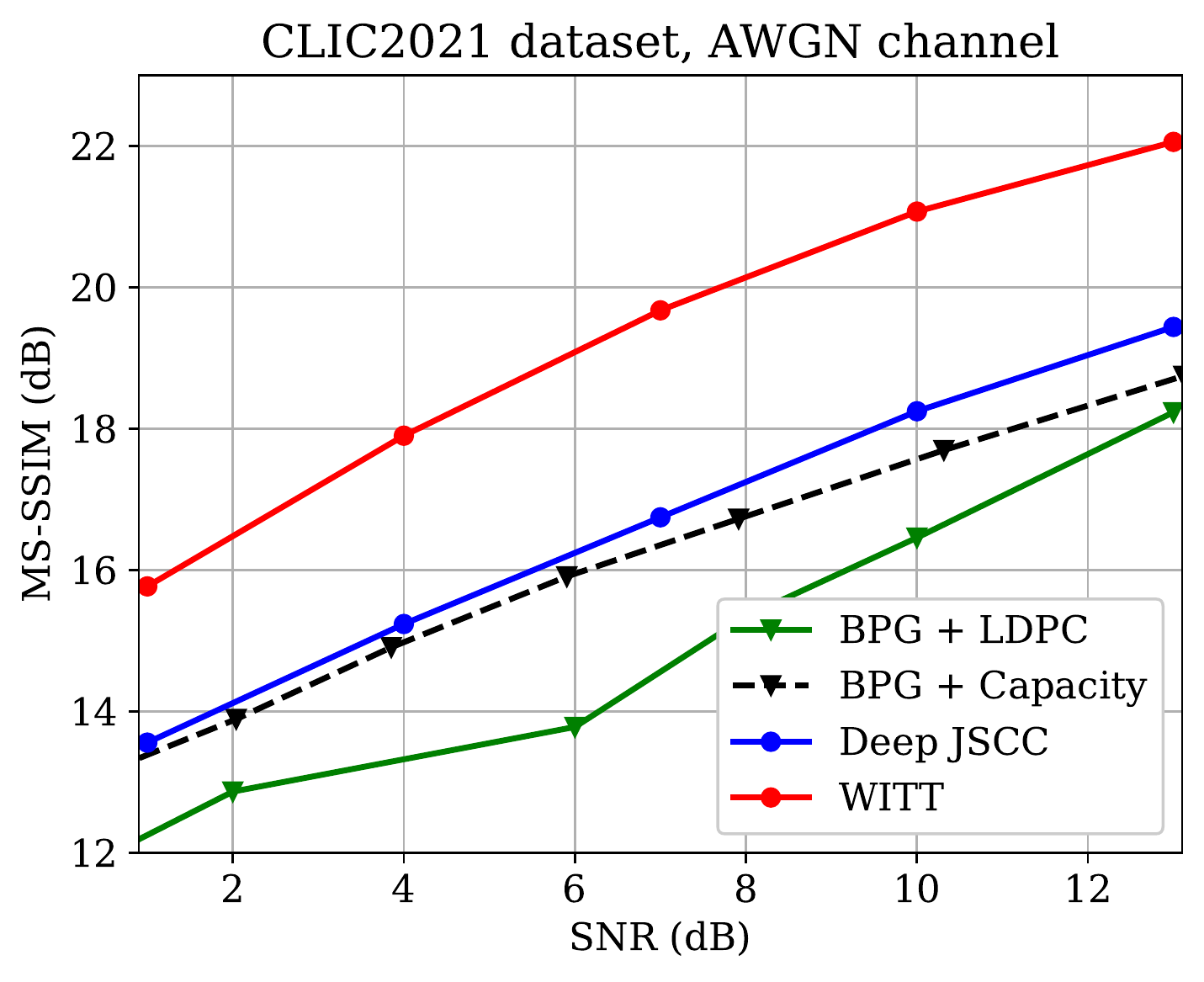}}
\hspace{-.10in}
\vspace{-2em}
\caption{MS-SSIM performance versus the SNR at the AWGN channel, and the average CBR is set to $1/16$.}\label{Fig5}
\vspace{-2.2em}
\end{center}
\end{figure}
\vspace{-0.5em}

\section{EXPERIMENTAL RESULTS}\label{section:experimental results}
\vspace{-0.5em}
\subsection{Experimental Setup}
 \noindent \textbf{Datasets: }We train and evaluate the proposed WITT scheme on image datasets with different resolutions from 32x32 upto 2K. For low-resolution images, we use the CIFAR10\cite{CIFAR10} dataset for training and testing. For high-resolution images, we choose DIV2K\cite{8014884} dataset for training, and use the Kodak\cite{Kodak} dataset and the CLIC2021\cite{CLIC2021} testset for testing. During training, images are randomly cropped into $256 \times 256$ patches. \\
\textbf{Comparsion Schemes: }We compare our WITT scheme with the CNN-based deep JSCC scheme\cite{DJSCC} and classical separation-based source and channel coding schemes. Specifically, we employ the BPG\cite{BPG} codec for compression combined with 5G LDPC codes\cite{LDPC_5G} for channel coding (marked as ``BPG + LDPC''). Here, we considered 5G LDPC codes with a block length of 6144 bits for different coding rates and quadrature amplitude modulations (QAM). Moreover, the ideal capacity-achieving channel code is also considered during evaluation (marked as ``BPG + Capacity'').\\
\textbf{Evaluation Metrics: } We qualify the performance of the proposed scheme using both the widely used pixel-wise metric PSNR and the perceptual metric MS-SSIM\cite{wang2003multiscale}. For PSNR, we optimized our model by the mean square error (MSE) loss function. For MS-SSIM, the loss function is 1 $-$ MS-SSIM.\\
\textbf{Training Details: }The number of stages in WITT varies with training image resolution. For low-resolution images, we use 2 stages with $[N_1, N_2]=[2, 4]$, $[C_1, C_2]=[128, 256]$, and the window size is set to 2. For large-resolution images, we use 4 stages $[N_1, N_2, N_3, N_4]=[1, 1, 2, 6]$, $[C_1, C_2, C_3, C_4]=[128, 192, 256, 320]$, and the window size is set to 8. During the training process, we first train other parameters except for the Channel ModNet over the wireless channel. Then, the whole proposed model is trained with Channel ModNet. We exploit the Adam optimizer with a learning rate of $1 \times 10^{-4}$, and the batch size is set to 128 and 16 for CIFAR10 dataset and DIV2K dataset, respectively. The WITT model is trained under the channel with a uniform distribution of SNR$_{train}$ from 1dB to 13 dB.

\vspace{-0.5em}
\subsection{Result Analysis}

Fig. \ref{Fig3}(a) $\sim$ \ref{Fig3}(c) show the PSNR performance versus SNR over the AWGN channel, and Fig. \ref{Fig3}(d) $\sim$ \ref{Fig3}(f) present the Rayleigh fast fading channel. For the WITT scheme, a single model can cover a range of SNR from 1dB to 13dB. For the ``BPG + LDPC'' scheme, according to adaptive modulation and coding (AMC) standard\cite{3gpp}, we choose the best-performing configuration of coding rate and modulation (the green dashed lines) under each specific SNR and plot the envelope. Compared to the CNN-based deep JSCC scheme, we achieve much better performance for all SNRs. Due to the enhanced model capacity by incorporating Transformers, it can be seen that the performance gap increases with the growth of image resolution. For the CIFAR10 dataset, WITT and deep JSCC scheme significantly outperform the ``BPG + LDPC'' and ``BPG + Capacity''. However, for high-resolution images, the performance of CNN-based deep JSCC degrades a lot and falls behind to the separation-based scheme. Our proposed maintains a considerable performance, especially in the low SNR regions.

Fig. \ref{Fig4}(a) $\sim$ \ref{Fig4}(c) demonstrate the PSNR performance versus the CBR over the AWGN channel, and Fig. \ref{Fig4}(d) $\sim$ \ref{Fig4}(f) show the Rayleigh fast fading channel. For the CIFAR10 dataset, our proposed model can generally outperform deep JSCC for all CBRs. Meanwhile, our model achieves considerable gains compared to the existing classical separation-based schemes, especially on the Rayleigh channel. For high-resolution image datasets, our proposed model outperforms the CNN-based deep JSCC scheme. Compared to ``BPG + LDPC'', our WITT model achieves comparable or better performance and coding gain. Moreover, WITT cannot provide comparable coding gain as that of the ``BPG + Capacity'' scheme, i.e., the slope of the performance curve slows down with the increase of SNR. Nevertheless, the performance of WITT approaches the ``BPG + Capacity'' in the low CBR regions and obviously improves compared to CNN-based deep JSCC.

Besides, to more comprehensively evaluate the performance of our model, we also train and test our model on the MS-SSIM metric. MS-SSIM is a multi-scale perceptual metric that approximates human visual perception well. Fig. \ref{Fig5} shows the performance versus the SNR at the AWGN channel with an average CBR $= 1/16$. For more intuitive observation and comparison, it is converted into the form of $\text{dB}$ and the formula is $\text{MS-SSIM(dB)} = -10\text{log(1}-\text{MS-SSIM)}$. Results indicate that the proposed WITT model can outperform other competitors by a large margin. Compared to the PSNR results in Fig. \ref{Fig3}, we can find that classical image transmission series are inferior to the learning-based WITT because classical image compression is designed to be optimized for squared error with hand-crafted constraints. Fig. \ref{Fig6} visualizes the reconstructions. It can be observed that WITT can achieve better visual quality with the same or lower channel bandwidth cost. More specifically, it avoids block artifacts and produces higher fidelity textures and details.

Table \ref{Table1} lists the inference time, FLOPs, and model size (\#param.) for WITT and ADJSCC \cite{ADJSCC} on Kodak dataset with a batch size of 1. All experiments are carried out using a Linux server with a single RTX 3090 GPU. Benefiting from the high-efficient window-based attention mechanism, WITT spends 2.5x lower floating point of operations (FLOPs). Despite its larger model size, WITT can provide better performance and run faster than the ADJSCC.

\begin{table}[t]
\renewcommand{\arraystretch}{1.3}
  \centering
  \small

  \caption{Inference speed and complexity comparison.}
\label{table1}

  \begin{tabular}{|m{1.5cm}|m{2cm}|m{1.5cm}|m{1.5cm}|}

    \Xhline{1pt}

    \centering {method} & \centering {inference time} & \centering {FLOPs} & \centering {\#param.}  \tabularnewline

     \hline

     \centering {WITT} & \centering $116${ms} & \centering $198${G} & \centering $28.2${M} \tabularnewline

     \hline

     \centering {ADJSCC} & \centering $155${ms} & \centering $511${G} & \centering $16.2${M}  \tabularnewline

     \Xhline{1pt}

  \end{tabular}
  \vspace{-1em}
  \label{Table1}
\end{table}

\section{Conclusion}\label{sec:conclusion}
\vspace{-0.5em}
In this paper, we have proposed a high-efficiency scheme named WITT to improve the performance of wireless image transmission.
The WITT framework is built upon the Swin Transformer to extract long-term hierarchical image representation. To deal with various channel conditions, we have further proposed the Channel ModNet to rescale the representations according to channel states automatically. Results have demonstrated that our proposed method outperforms the CNN-based deep JSCC scheme and the classical separated-based schemes.

\vfill
\pagebreak

\bibliographystyle{IEEEbib}
\bibliography{main}

\end{document}